# DALC: Distributed Automatic LSTM Customization for Fine-Grained Traffic Speed Prediction

Ming-Chang Lee[1] and Jia-Chun Lin[2]

*Department of Information Security and Communication Technology, Norwegian University of Science and Technology,*
*Ametyst-bygget, 2815 Gjøvik, Norway*

[1]*ming-chang.lee@ntnu.no*
[2]*jia-chun.lin@ntnu.no*

# DALC: Distributed Automatic LSTM Customization for Fine-Grained Traffic Speed Prediction

Ming-Chang Lee[1] and Jia-Chun Lin[2]
Department of Information Security and Communication Technology, Norwegian University of Science and Technology, 2815 Gjøvik, Norway
[1] ming-chang.lee@ntnu.no
[2] jia-chun.lin@ntnu.no

**Abstract.** Over the past decade, several approaches have been introduced for short-term traffic prediction. However, providing fine-grained traffic prediction for large-scale transportation networks where numerous detectors are geographically deployed to collect traffic data is still an open issue. To address this issue, in this paper, we formulate the problem of customizing an LSTM model for a single detector into a finite Markov decision process and then introduce an Automatic LSTM Customization (ALC) algorithm to automatically customize an LSTM model for a single detector such that the corresponding prediction accuracy can be as satisfactory as possible and the time consumption can be as low as possible. Based on the ALC algorithm, we introduce a distributed approach called Distributed Automatic LSTM Customization (DALC) to customize an LSTM model for every detector in large-scale transportation networks. Our experiment demonstrates that the DALC provides higher prediction accuracy than several approaches provided by Apache Spark MLlib.

## 1 Introduction

In the past decade, several approaches for short-term traffic prediction have been proposed. They can be classified into parametric approaches and nonparametric approaches. The autoregressive integrated moving average (ARIMA) model is a widely used parametric approach [1], in which the model structure is predefined. The nonparametric approaches include k-nearest neighbors method, artificial neural network, recurrent neural network (RNN), etc. As a type of RNN, long short-term memory (LSTM) [2] is superior in predicting time series problem with long temporal dependency such as traffic prediction. Prior study [3][4][5][22] have demonstrated that LSTM provides satisfactory prediction accuracy. However, to our knowledge, none of existing LSTM-based prediction methods is designed to provide fine-grained traffic prediction for large-scale transportation networks where numerous detectors are geographically deployed to collect traffic data.

The success of LSTM depends on choosing an appropriate hyperparameter configuration, including the number of hidden layers and the number of epoch [6], since the configuration determines if LSTM can achieve satisfactory prediction accuracy or not. However, determining such a configuration is usually done manually. Each time a different configuration is used for training an LSTM model, and many times of retrainings might be required until the LSTM model provides satisfactory prediction performance. This process might be time consuming and energy-inefficient, and such a process will be even longer when providing the above-mentioned fine-grained traffic prediction for large-scale transportation networks.

To address the above issues, in this paper, we formulate the problem of customizing an LSTM model with an appropriate hyperparameter configuration for a single detector into a Markov decision process and then employ Value Iteration [7] to suggest the policy that consumes the least expected training time. We then incorporate the policy into an automatic LSTM customization (ALC) algorithm and further take prediction accuracy into account to automatically customize an appropriate LSTM model for a single detector. More specifically, ALC will keep training the LSTM model by preferentially following the policy suggested by Value Iteration until the prediction accuracy of the LSTM model reaches a predefined threshold or until the prediction accuracy cannot be further improved by all possible choices.

In order to provide fine-grained traffic speed prediction, we propose that each detector-period combination (DPC), i.e., each detector in a different time period, should have its own LSTM model. In addition, to effectively customize LSTM models for all DPCs in large-scale transportation networks, we introduce a distributed approach based on the ALC algorithm, named DALC. Note that the first letter D stands for "distributed". In the DALC, each DPC will have its own LSTM model, and the jobs for customizing LSTM models for all DPCs will be executed by a set of computation nodes in a parallel manner.

To demonstrate the effectiveness of DALC, we conduct an experiment to compare DALC with several distributed machine learning approaches provided by Apache Spark MLlib [8]. The results show that DALC provides the best prediction accuracy and is able to achieve fine-grained traffic speed prediction for large-scale transportation networks in a distributed and parallel manner.

The rest of the paper is organized as follows: Sections 2 and 3 describe the background of LSTM and related work, respectively. In Section 4, we introduce the details of ALC and DALC. Section 5 presents the experiment result. In Section 6, we conclude this paper and outline future work.

## 2   LSTM

LSTM [2] is a special type of RNNs with ability to learn long-term dependencies and model temporal sequences. The architecture of LSTM is similar to that of RNN except that the nonlinear units in the hidden layers are replaced by memory blocks. Each memory block contains one or more self-connected memory cells to store internal state. Each memory block also contains three multiplicative units (input, output and forget gates) to manage cell state and output using activation functions. These features enable LSTM to preserve information in the memory block over long time lags.

In order to optimize the prediction performance of LSTM, it is essential to choose appropriate hyperparameters, including the number of hidden layers, the number of hidden units, the number of epochs (Note that an epoch is defined as a complete pass through a given dataset [6]), learning rate, activation function, etc. Determining the above hyperparameters often depends on a trial-and-error approach and lots of practices and experiences. In this paper, we focus on determining a configuration consisting of two hyperparameters, i.e., the number of hidden layers and the number of epochs, since these two hyperparameters are influential on determining both training time and prediction accuracy. Our goal is to automatically customize an LSTM model with an appropriate hyperparameter configuration for a detector such that the prediction accuracy can be as satisfactory as possible and the corresponding time consumption can be as low as possible.

## 3     Related work

Existing traffic prediction approaches can be classified into two categories: parametric approaches and nonparametric approaches.

Parametric approaches are also called model-based methods in which the model structure has to be determined in advance based on some theoretical assumptions, and the model parameters can be derived with empirical data. The autoregressive integrated moving average (ARIMA) model is a widely used parametric approach [9], with which Ahmed and Cook [1] predicted short-term freeway traffic flow and Hamed et al. [10] forecasted traffic volume in urban arterial roads. Many ARIMA-based approaches were then developed to enhance prediction accuracy, including Kohonen-ARIMA [11] and seasonal ARIMA [12].

Different from parametric approaches, nonparametric approaches do not require a predefined model structure. Typical examples of nonparametric approaches include k-nearest neighbors (k-NN), artificial neural network (ANN), RNN, hybrid approaches, etc. In year 1991, the k-NN method was used by Davis and Nihan [13] to forecast freeway traffic. After that, several variants of the k-NN method were introduced for traffic prediction. For instances, Bustillos et al. [14] proposed a travel time prediction model based on n-curve and k-NN methods.

Lv et al. [9] proposed a deep learning approach with a stacked autoencoder model to learn generic traffic flow features for traffic flow prediction. The greedy layerwise unsupervised learning algorithm is applied to pre-train the deep network, and then a fine-tuning process is used to update the parameters of the model so as to improve prediction accuracy. Ma et al. [3] employed LSTM to forecast traffic speed using remote microwave sensor data. Their experiment results compared with other recurrent neural networks (including Elman NN, Time-delayed NN, Nonlinear Autoregressive NN, support vector machine, ARIMA, and the Kalman Filter approach) show that LSTM provides superior prediction accuracy and stability.

Different from all the above work, in this paper, we focus on providing fine-grained traffic speed prediction for large-scale transportation networks in a distributed and parallel manner. Customizing an LSTM for a single detector is automatically done by the proposed ALC algorithm. In addition, to effectively customize LSTMs for the enormous number of detectors in the target large-scale transportation networks, we introduce a distributed approach based on the ALC algorithm.

## 4 LSTM customization for a single detector

In this section, we introduce how to convert the LSTM customization problem for a single detector into a finite Markov decision process (MDP), and then present the ALC algorithm to achieve automatic customization.

### 4.1 Markov decision process formulation

As mentioned earlier, this paper focuses on customizing an LSTM model for a detector in terms of two-hyperparameter configuration: hidden layers and the number of epoch. For every detector, its LSTM can have up to $n$ hidden layers and the maximum allowed training for every different configuration is $k$ epochs, where $n \geq 1$ and $k \gg 1$. Fig. 1 illustrates the state transition graph for the LSTM customization problem. Each state $s$ is a large oval labelled by the number of hidden layers, and the number of epochs, except the start state which is labelled by *start*. We define an LSTM model under configuration $\langle h, j \cdot e \rangle$ as state $s_{h,j \cdot e}$, implying that the LSTM model has been trained with the configuration of $h$ hidden layers and $j \cdot e$ epochs, where $h \leq n$, $e$ is a fixed integer number (e.g., 100), and $j = 1, 2, \ldots, k/e$. For instance, when the state is $s_{1,e}$, it means that the LSTM model has been trained with configuration $\langle 1, e \rangle$. Note that the number of epochs is assumed to start from $e$, regardless of the value of $h$.

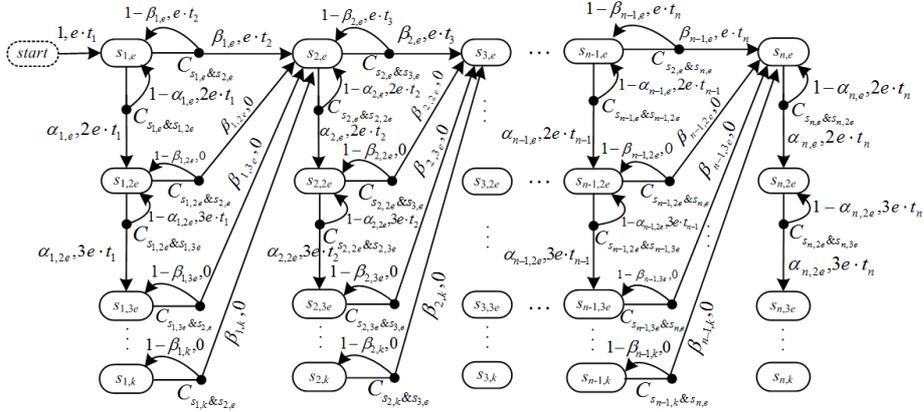

**Fig. 1.** The state transition graph of customizing an LSTM model for a detector.

We define the action set for state $s$ to be $A_s$. Given the current state $s_{h,j \cdot e}$, two actions denoted by small solid circles in Fig. 1 are possible: $C_{s_{h,j \cdot e} \& s_{h,(j+1) \cdot e}}$ and $C_{s_{h,j \cdot e} \& s_{h+1,e}}$, i.e., $A_{s_{h,j \cdot e}} = \{ C_{s_{h,j \cdot e} \& s_{h,(j+1) \cdot e}}, C_{s_{h,j \cdot e} \& s_{h+1,e}} \}$. The former action is to compare the LSTM model under configuration $\langle h, j \cdot e \rangle$ with the LSTM model under configuration $\langle h, (j + 1) \cdot e \rangle$. If the latter LSTM provides better prediction accuracy, the state will transit to $s_{h,(j+1) \cdot e}$. Otherwise, the state will remain the same, i.e., $s_{h,j \cdot e}$. The latter action $C_{s_{h,j \cdot e} \& s_{h+1,e}}$ is to compare if the prediction accuracy of the LSTM model can be improved by changing the configuration from $\langle h, j \cdot e \rangle$ to $\langle h + 1, e \rangle$, i.e., adding one more hidden layer with the initial number of epochs. If the answer is true, the state will transit to $s_{h+1,e}$. Otherwise, the state will still be $s_{h,j \cdot e}$.

**Table 1.** All possible values for state transition probability and time consumption.

| s | a | s' | $P(s'\|s,a)$ | $T(s,a,s')$ |
|---|---|---|---|---|
| $s_{h,j \cdot e}$ | $C_{s_{h,j \cdot e} \& s_{h,(j+1) \cdot e}}$ | $s_{h,(j+1) \cdot e}$ | $\alpha_{h,j \cdot e}$ | $(j+1) \cdot e \cdot t_h$ |
|  |  | $s_{h,j \cdot e}$ | $1 - \alpha_{h,j \cdot e}$ | $(j+1) \cdot e \cdot t_h$ |
|  | $C_{s_{h,j \cdot e} \& s_{h+1,e}}$ | $s_{h+1,e}$ | $\beta_{h,j \cdot e}$ | $e \cdot t_{h+1}$ |
|  |  | $s_{h,j \cdot e}$ | $1 - \beta_{h,j \cdot e}$ | $e \cdot t_{h+1}$ |

Based on the chosen action $a$ in state $s$, the state transition probability function for the next state $s'$ is denoted by $P(s'|s,a)$. Let $T(s,a,s')$ be the time consumption incurred by taking action $a$ in state $s$ to transit to state $s'$. All possible values for $P(s'|s,a)$ and $T(s,a,s')$ are listed in Table 1. Taking action $C_{s_{h,j \cdot e} \& s_{h,(j+1) \cdot e}}$ in state $s_{h,j \cdot e}$ means that the LSTM model needs to be retrained with configuration $\langle h, (j+1) \cdot e \rangle$, i.e., $h$ hidden layers with $(j+1) \cdot e$ epochs. Hence, the corresponding time consumption is $(j+1) \cdot e \cdot t_h$ where $t_h$ is the time for executing an epoch when the number of hidden layers is $h$. On the other hand, taking action $C_{s_{h,j \cdot e} \& s_{h+1,e}}$ in state $s_{h,j \cdot e}$ means that the LSTM model needs to be retrained with configuration $\langle h+1, e \rangle$, i.e., $h+1$ hidden layers with $e$ epochs. Therefore, the corresponding time consumption is $e \cdot t_{h+1}$ where $t_{h+1}$ is the execution time per epoch when the number of hidden layers is $h+1$.

### 4.2 The ALC algorithm

To find an appropriate hyperparameter configuration for a detector such that the resulting LSTM model is able to provide satisfactory prediction accuracy with low time consumption, we propose the ALC algorithm based on the state transition graph shown in Fig. 1 and Value Iteration [7]. Value Iteration is an iterative method of computing an optimal MDP policy and its value. Let $Q_i(s,a)$ be the action-value function assuming there are $i$ steps to go from state $s$ by taking action $a$. Let $V_i(s)$ be the state-value function assuming there are $i$ steps to go from state $s$.

$Q_i(s,a) = \sum_{s'} P(s'|s,a) \cdot (T(s,a,s') + \gamma \cdot V_{i-1}(s'))$ for $i > 0$.

$V_i(s) = \min_a Q_i(s,a) = \min_a \sum_{s'} P(s'|s,a) \cdot (T(s,a,s') + \gamma \cdot V_{i-1}(s'))$ for $i > 0$.

where $\gamma$ is a discount rate, which equals to 1 in this paper so that all the costs can be accumulated as they are.

Fig. 2 shows the ALC algorithm. By starting with an arbitrary function $V_0$ (i.e., $i = 0$) and using the above two equations to get the functions for $i+1$ steps to go from the functions for $i$ steps to go (i.e., working backward), the ALC algorithm calculates $V_i(s)$ for each state $s$ and then checks if $|V_i(s) - V_{i-1}(s)|$ is larger than $\theta$ for all the states (see lines 3 to 7), where $\theta$ is a predefined threshold with a positive value. If the answer is yes, implying that the difference between the two expected time consumptions is more than we accept, the ALC algorithm terminates its searching. As line 9 shows, for each state $s$, the action leading to the least expected time consumption will be stored as $\pi(s)$, i.e., $\pi(s)$ is the action suggested by Value Iteration to take when the state is $s$.

---

**The ALC algorithm**
**Input:** The training data and testing data associated with a detector
**Output:** An LSTM model with an appropriate hyperparameter configuration for the detector
**Procedure:**

1:    Let $V_0(s) = 0$ for each state $s$;
2:    Let $i = 0$;
3:    **repeat**
4:      $i = i + 1$;
5:    **for** each state $s \in S$ {
6:      $V_i(s) = \min_a \sum_{s'} P(s'|s,a) \cdot (T(s,a,s') + V_{i-1}(s'));$}
7:    **until** $\forall s \ |V_i(s) - V_{i-1}(s)| > \theta$
8:    **for** each state $s \in S$ {
9:      $\pi(s) = \arg\min_a \sum_{s'} P(s'|s,a) \cdot (T(s,a,s') + V_{i-1}(s'));$}
10:   Let $E_{now} = 0$ and $E_{new} = 0$; //They are used to store the AARE of current LSTM and new LSTM
11:   Let $h = 1, j = 1$, and $f =$ false;
12:   Use configuration $\langle 1, e \rangle$ to train an LSTM model;
13:   $E_{now} =$ the AARE of this LSTM model;
14:   **if** $E_{now} \leq \delta$ { // $\delta$ is a predefined threshold.
15:    Output the LSTM model under configuration $\langle 1, e \rangle$; $f =$ true;}
16:   **else** {
17:    **while** $f =$ false & $j \leq k/e$ & $h \leq n$ {
18:     **if** $\pi(s_{h,j\cdot e}) = C_{s_{h,j\cdot e} \& s_{h,(j+1)\cdot e}}$ { //Follow the action suggested at line 9.
19:      Use configuration $\langle h, (j + 1) \cdot e \rangle$ to retrain the LSTM model;
20:      $E_{new} =$ the AARE of the new LSTM model;
21:      **if** $E_{new} < E_{now}$ { //The suggested action can lower current prediction error.
22:       **if** $E_{new} \leq \delta$ { Output the LSTM model under $\langle h, (j + 1) \cdot e \rangle$; $f =$ true;}
23:       **else** { $j = j + 1$; $E_{now} = E_{new}$; }}; // To continue by increasing number of epochs.
24:      **else** {//The suggested action cannot lower the prediction error, so try the other action.
25:       Use configuration $\langle h + 1, e \rangle$ to retrain the LSTM model;
26:       $E_{new} =$ the AARE of the new LSTM model;
27:       **if** $E_{new} < E_{now}$ {
28:        **if** $E_{new} \leq \delta$ { Output the LSTM model under $\langle h + 1, e \rangle$; $f =$ true; }
29:        **else** { $h = h + 1$; $j = 1$; $E_{now} = E_{new}$; }}
30:       **else** { //Both actions cannot improve the AAREs of current LSTM and new LSTM.
31:        Output the LSTM model under $\langle h, j \cdot e \rangle$; $f =$ true;}}}
32:     **else** { //It means that the suggested action is $C_{s_{h,j\cdot e} \& s_{h+1,e}}$.
33:      Use configuration $\langle h + 1, e \rangle$ to retrain the LSTM model;
34:      $E_{new} =$ the AARE of the new LSTM model;
35:      **if** $E_{new} < E_{now}$ { //The suggested action can lower current prediction error.
36:       **if** $E_{new} \leq \delta$ { Output the LSTM model under $\langle h + 1, e \rangle$; $f =$ true;}
37:       **else** { $h = h + 1$; $j = 1$; $E_{now} = E_{new}$; }}
38:      **else** {//The suggested action cannot lower the prediction error, so try the other action.
39:       Use configuration $\langle h, (j + 1) \cdot e \rangle$ to retrain the LSTM model;
40:       $E_{new} =$ the AARE of the new LSTM model;
41:       **if** $E_{new} < E_{now}$ {
42:        **if** $E_{new} \leq \delta$ { Output the LSTM model under $\langle h, (j + 1) \cdot e \rangle$; $f =$ true;}
43:        **else** { $j = j + 1$; $E_{now} = E_{new}$; }}
44:       **else** { //Both actions cannot improve the AAREs of current LSTM and new LSTM.
45:        Output the LSTM model under $\langle h, j \cdot e \rangle$; $f =$ true; }}}

**Fig. 2.** The ALC algorithm.

Following all suggested actions can lead the total time consumption to the minimum, but it does not guarantee that the resulting configuration can achieve satisfactory prediction accuracy. On the other hand, keep searching for a configuration and use it to retraining the LSTM might be able to keep enhancing the prediction accuracy, but it might take a very long time. To avoid unnecessary time consumption, the ALC algorithm keeps searching for configurations that can enhance prediction accuracy and terminates when the LSTM under a configuration provides satisfactory prediction accuracy or when the prediction accuracy cannot be improved by all possible choices. The detailed process is as follows: The algorithm first uses configuration $\langle 1, e \rangle$, i.e., one hidden layer with $e$ epochs, to train an LSTM model (see line 12). If the average absolute relative error (AARE) of the LSTM model (which is calculated based on Equation 1) is less than a predefined threshold $\delta$ (i.e., line 14), implying that the prediction accuracy is satisfactory, then the ALC algorithm outputs this LSTM model to be the LSTM model of the detector and sets $f$ to be true so as to terminate the search process. Otherwise, the ALC algorithm takes the action suggested by Value Iteration for state $s_{h,j\cdot e}$, i.e., $\pi(s_{h,j\cdot e})$. Note that $f$ is a boolean variable indicating if the desired LSTM model is derived or not.

If $\pi(s_{h,j\cdot e})$ is $C_{s_{h,j\cdot e}\&s_{h,(j+1)\cdot e}}$ (see line 18), the algorithm compares the LSTM model under configuration $\langle h, j \cdot e \rangle$ with the LSTM model under configuration $\langle h, (j+1) \cdot e \rangle$, implying that the algorithm must retrain the LSTM model with $\langle h, (j+1) \cdot e \rangle$. If this new LSTM model provides a lower AARE than the original one (see line 21), the algorithm further checks if the AARE of this new LSTM model is lower or equal to threshold $\delta$. As line 22 shows, if the answer is yes, meaning that the prediction accuracy is satisfactory, this LSTM model is outputted to be the LSTM model of the detector. Otherwise, the algorithm tries another configuration by increasing $j$ by one (see line 23) to attempt achieving satisfactory prediction accuracy. The algorithm will go back to line 17 and to see if it can proceed.

However, as line 24 shows, if the LSTM model under configuration $\langle h, (j+1) \cdot e \rangle$ is worse than the LSTM model under configuration $\langle h, j \cdot e \rangle$, implying that the action suggested by Value Iteration is unable to enhance the prediction accuracy, the algorithm will take the other action, i.e., $C_{s_{h,j\cdot e}\&s_{(h+1),e}}$. In this case, the LSTM model will be retrained with configuration $\langle h+1, e \rangle$ (see line 25). If the prediction accuracy of this new LSTM model is satisfactory, it will be outputted (see line 28). In the case that this new LSTM model is better than the previous one but its AARE is still not low than $\delta$, the algorithm will try another configuration by increasing $h$ by one and setting $j$ to be one (see line 29). The algorithm will go back to line 17 and to see if it can proceed. It might be possible that the LSTM model under $\langle h+1, e \rangle$ is worse than that under $\langle h, j \cdot e \rangle$ (see line 30), it means that neither taking action $C_{s_{h,j\cdot e}\&s_{h,(j+1)\cdot e}}$ nor $C_{s_{h,j\cdot e}\&s_{(h+1),e}}$ can enhance the prediction accuracy. In this case, the algorithm outputs the LSTM model under configuration $\langle h, j \cdot e \rangle$ to avoid unnecessary time consumption.

The algorithm will follow a similar procedure as mentioned above to customize an appropriate LSTM model for the detector when $\pi(s_{h,j\cdot e})$ is $C_{s_{h,j\cdot e}\&s_{h+1,e}}$ (see lines 32 to 45).

## 4.3 DALC

In this section, we introduce how to customize LSTMs for all detectors in large-scale transportation networks based on the ALC algorithm. This paper focuses on predicting traffic speed in two specific periods on weekdays. One is from 4 am to 10 am. The other is from 2 pm to 8 pm. The reason we choose these two periods is that they cover peak commute hours, which might significantly affect traffic speed.

Due to the dynamic nature of large-scale transportation networks, detectors deployed in different places might have diverse traffic-speed patterns in the two abovementioned periods. To demonstrate this, we choose ten detectors deployed between mile 1.14 and mile 14.4 on freeway I5-N in California [21] to compare their traffic-speed patterns in the AM period of a typical weekday. As illustrated in Fig. 3, not all of their patterns are identical. Hence, we propose that each detector should have its own LSTM model in order to achieve fine-grained traffic speed prediction.

Furthermore, for any single detector, it is also possible that its traffic-speed patterns in these two periods are completely different from each other. According to our observation, we found that many detectors deployed on freeway I5-N have this phenomenon. For instance, the traffic speed collected by detector 1114190 (which is one of the detectors in Fig. 3) for five consecutive weekdays (from Oct. 16[th] 2017 to Oct. 20[th] 2017) illustrated in Fig. 4 shows that the pattern in the AM period is totally different from that in the PM period. Therefore, we propose that each detector in each of the two periods should have its own LSTM model in order to achieve our goal. In other words, the total number of LSTM models will be $2m$ if $m$ is the total number of the detectors in large-scale transportation networks.

To effectively customize LSTMs for each of the $2m$ detector-period combinations (DPCs for short) in parallel, we extend ALC in a distributed and parallel manner and call the distributed approach DALC. DALC utilizes a set of computation nodes to share the workload of customizations. As long as a computation node is available, DALC requests it to customize a LSTM model for a DPC. In this way, LSTM customization for all the $2m$ DPCs can be conducted in parallel.

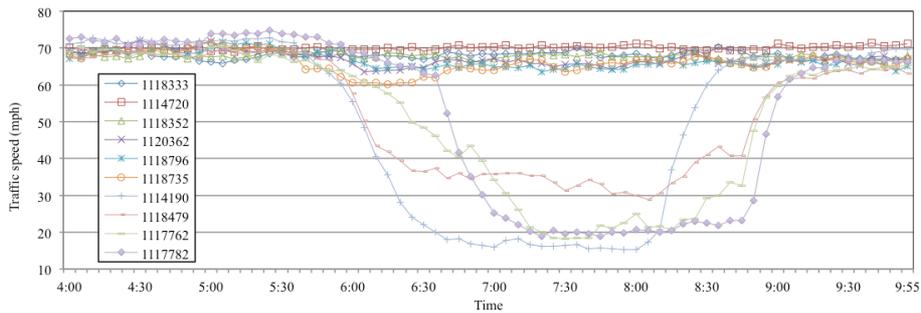

**Fig. 3.** The traffic speed collected by ten detectors on freeway I5-N in the AM period of Oct. 16[th], 2017.

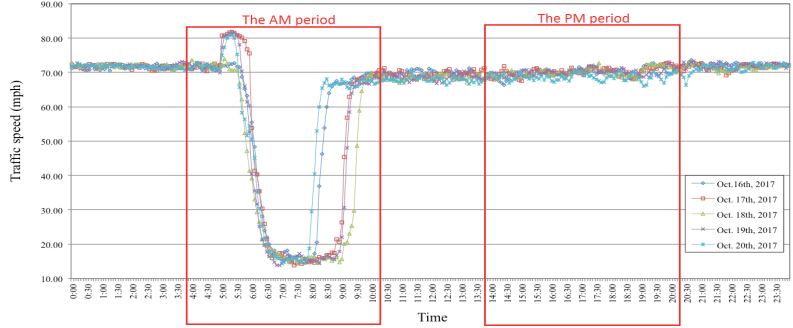

**Fig. 4.** The traffic speed collected by detector 1114190 deployed on freeway I5-N for five consecutive weekdays.

## 5   Experiment results

We validated the prediction accuracy of our proposed approach in comparison with five distributed machine learning approaches provided by Apache Spark MLlib [8], including Linear Regression (LR), Generalized Linear Regression (GLR), Decision Tree Regression (DTR), Gradient Boosted Tree Regressor (GBTR), and Random Forest Regressor (RFR). All the six approaches are applied to the traffic data collected by the California Department of Transportation Performance Measurement System [21], which is a consolidated database of traffic data collected at 5-minute intervals by each detector placed on state highways throughout California. In this paper, we concentrate on predicting traffic speed on freeway I5-N.

**Table 2.** The average training time per epoch given different number of hidden layers.

| Number of hidden layers | Average training time per epoch |
|---|---|
| 1 | $t_1$= 2.214 sec |
| 2 | $t_2$= 3.311 sec |
| 3 | $t_3$= 4.728 sec |
| 4 | $t_4$= 5.547 sec |
| 5 | $t_5$= 6.754 sec |

We established a private cluster using Hadoop YARN 2.2.0 [15] and Apache Spark 2.0.1 [16]. The reason we chose Hadoop YARN is that it is an open-source software framework with high scalability, efficiency, and flexibility for processing high volume of dataset [17][18]. This cluster consists of one master node and 30 slave nodes. Each node ran Ubuntu 12.04.1 LTS with 2 CPU Cores, 2GB of RAM, and 100 GB of storage. To guarantee a fair comparison, no other job or work was executed when each of the abovementioned approaches runs on the cluster. When the five MLlib approaches were employed, they utilized current traffic flow to predict future traffic speed in 5-minute intervals. For DALC, we used DL4J [19] to implement the corresponding LSTM and adopted the default suggested values for all hyperparameters [19], except the two parameters considered in this paper, i.e., the number of hidden layers and the number of epochs. Recall that the average training time for each epoch under different number of hidden layers is required. This information is shown in Table 2 after we ran some experiments on the cluster. We can see that $t_1 < t_2 < t_3 < t_4 < t_5$, implying that the training time for each epoch increases as the number of hidden layers increases. By following the suggestion from [20] to achieve highly accurate prediction capability, the threshold $\delta$ used in the ALC algorithm is 0.05 for our approaches.

To extensively measure and compare the effectiveness of all the approaches, one widely used performance metric, i.e., average absolute relative error (AARE) is employed, and it is defined as follows:

$$\text{AARE} = \frac{1}{W} \cdot \sum_{w=1}^{W} \frac{|\tau_w - \widehat{\tau_w}|}{\tau_w} \qquad (1)$$

where $W$ is the total number of data samples for comparison, $w$ is the index of time point, $\tau_w$ is the observed traffic speed at time point $w$, and $\widehat{\tau_w}$ is the predicted traffic speed at time point $w$.

In this experiment, we selected 60 detectors deployed on freeway I5-N ranging from mile 0 to mile 150.35 to be our targets. Recall that this paper focuses on providing traffic speed prediction for every detector in two specific AM and PM periods. Therefore, there are 120 DPCs (which stands for detector-period combinations) for the 60 detectors. For each DPC, we chose its traffic-speed data in the corresponding period from five weekdays (from Oct. 16th, 2017 to Oct. 20th, 2017) to be the training data of all the approaches, and chose its traffic-speed data in the corresponding period from the next three weekdays (i.e., Oct. 23th, 2017 to Oct. 25th, 2017) to be the testing data of all the approaches.

Fig. 5 illustrates the average AARE results of all approaches for the 120 DPCs. It is clear that the DALC approach outperforms the rest approaches. When DALC was employed, the average AARE values are all less than 0.04 with small standard deviation (see Fig. 5). However, when the rest five approaches were tested, the corresponding average AARE values are between 0.12 and 0.17 with significant standard deviation, implying that these five approaches provide offer poor prediction accuracy for the DPCs. In other words, they could not guarantee good prediction accuracy for all the DPCs.

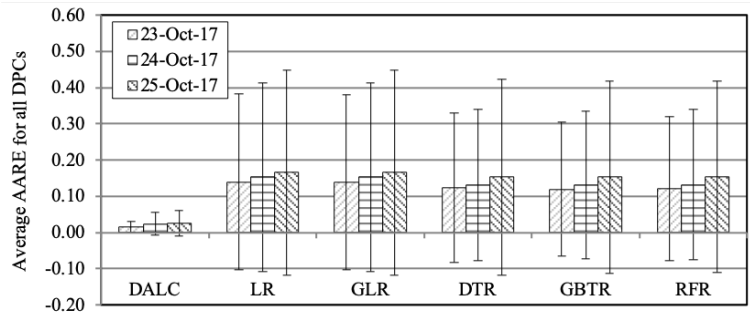

**Fig. 5.** The average AARE results of different approaches for all the 120 DPCs.

## 6 Conclusion and future work

In this paper, we have introduced the ALC algorithm to achieve automatic LSTM customization for a single detector by automatically configure the number of hidden layers and the number of epochs. Due to the diverse traffic patterns collected by detectors, we proposed to customize one LSTM model for each detector in a different time period (i.e., DPC). Furthermore, to effectively customize LSTMs for tremendous DPCs in large-scale transportation networks, we have introduced DALC to perform all the customization jobs in a distributed and parallel way. The experimental result based on real traffic data on freeway I5-N in California have demonstrated the outstanding prediction accuracy of DALC as compared with another five approaches provided by Apache Spark MLlib.

In our future work, instead of customizing one LSTM for every single DPC, we would like to cluster DPCs into groups if they all observe a similar traffic pattern and customize one LSTM model for each group so as to speed up LSTM customization for the entire large-scale transportation networks.

**Acknowledgments.** This work was supported by the project eX$^3$ - *Experimental Infrastructure for Exploration of Exascale Computing* funded by the Research Council of Norway under contract 270053 and the scholarship under project number 80430060 supported by Norwegian University of Science and Technology.